\documentclass[11pt,oneside,a4paper]{article}
\usepackage{authblk}
\usepackage{enumerate}
\usepackage{amsmath}
\usepackage{siunitx}
\usepackage{amssymb}
\usepackage{amsfonts}
\usepackage{graphicx}
\usepackage{subfigure}
\usepackage[linesnumbered,ruled,vlined]{algorithm2e}
\begin{document}

\title{Algorithms for an Efficient Tensor Biclustering}%\thanks{Supported by organization x.}}
\date{}
\author[1, 2]{A. D. Faneva}
\author[2]{M. Lebbah}
\author[2]{H. Azzag}
\author[2]{G. Beck}

\affil[1]{African Institute for Mathematical Sciences (AIMS), Km2 Route de Joal, Centre IRD, Mbour, BP 1418, Senegal}

\affil[2]{Computer Science Laboratory of Paris North (LIPN, CNRS UMR 7030), University of Paris 13, F-93430 Villetaneuse France}

\maketitle              % typeset the header of the contribution

\begin{abstract}
Consider a data set collected by \emph{(individuals-features)} pairs in different times. It can be represented as a tensor of three dimensions \emph{(Individuals, features and times)}. The tensor biclustering problem computes a subset of individuals and a subset of features whose signal trajectories over time lie in a low-dimensional subspace, modeling similarity among the signal trajectories while allowing different scalings across different individuals or different features. This approach are based on spectral decomposition in order to build the desired biclusters. We evaluate the quality of the results from each algorithms with both synthetic and real data set.

\providecommand{\keywords}[1]{\textbf{\textit{Index terms---}} #1}


\end{abstract}
	\section{Introduction}
Clustering analysis has become a fundamental tool in statistics and machine learning. Many clustering algorithms have been developed with the general
idea of seeking groups among different individuals in all space of features. 
%Inspired
%by the concept of direct clustering, 
Biclustering consists of simultaneous partitioning of a set of observations and a set of their features into subsets often called bicluster.
Consequently, a subset of rows exhibiting significant coherence within a subset of columns in the matrix can be extracted, which corresponds to a specific coherent pattern \cite{bou,chr}.
Nowadays, there is a new type of data collection, in which  we may collect data by  \emph{individual-feature} pair at multiple times. The variation of a couple \emph{(individual-feature)} at different instants is called trajectory. 
This data can be represented as a three dimensional object called  tensor  $\mathcal{T}\in \mathbb{R}^{n_1\times n_2\times m} $, where $n_1$ and $n_2$ are respectively the size of observations and features
%. One collects data for each \emph{(individual, feature)} pair 
at $m$ different times.
% 
%In this article, we propose several methods to solve this problem.
%
%Assume one have $n_1$ observations, $n_2$ features. We collect data for each \emph{(individual, feature)} pair at $m$ different times. The data can be represented as a three dimensional object $\mathcal{T}\in \mathbb{R}^{n_1\times n_2\times m} $, we call it \textbf{a tensor}. 
Tensor biclustering selects a subset of individual indices and a subset of features indices whose trajectories are highly correlated. Grouping those trajectories according to the correlation or similarity behaviour between them is useful in different area such as decision making, but it is still a very challenging topic in research.\\
\indent In  \cite{map}, the authors proposed different methods based on the spectral decomposition of matrix and the length of trajectory, although they provide a unique bicluster. 
%(see Fig.\ref{fig1}).  
Many tools on tensor manipulation already exist in literature  to solve this tensor biclustering problem \cite{jac,gas,gol,gud,fay}. Our algorithms are based on a spectral decomposition as proposed in \cite{map}.
This article is structured as follows. In Section~\ref{sec:theory}, we start by a brief summary of problem formulation. Section~\ref{sec:extension} introduces our algorithm extensions. Section~\ref{sec:exp} is related to the experiments.  We make some concluding remarks in Section~\ref{sec:conc}.

%\begin{figure}[!h]
%	\centering
%	%\subfigure[Mode-3 fibers $\mathcal{T}(i,j,:)$]{
%	\includegraphics[width=3.5cm,height=2.5cm]{tensor_2.png}
%	%	}
%	
%	\caption{Tensor with more than one blocks of bicluster. In this figure we have two blocks.}
%	\label{fig1}
%\end{figure}

%%%%
%\subsection*{Notation}

\section{Problem Formulation}
\label{sec:theory}
%We would like to recall some notations as used in 
We use the common notation where
%we have tried to remain as consistent as possible with terminology that would be familiar to applied mathematicians.
%
$\mathcal{T},\;\mathcal{X}$ and $\mathcal{Z}$ are used respectively to denote input, signal and noise tensors. For any set $J$, $|\bar{J}|$ denotes its cardinality.  $[n]$ denotes the set $\{1,2,\dots,n\}$. $|\bar{J}|=[n]-J$.  $\|x\|_{2}=(x^{t}x)^{1/2}$ is the second norm of the vector $x$.  $x\otimes y$ is the Kronecker product of two vectors $x$ and $y$.
We also use Matlab notation to denote the elements in tensor. Specifically, $\mathcal{T}(:,:,i)$, $\mathcal{T}(:,i,:)$ and $\mathcal{T}(i,:,:)$ are respectively the $i-th$ frontal, lateral and horizontal slice. $\mathcal{T}(:,i,j)$, $\mathcal{T}(i,:,j)$ and $\mathcal{T}(i,j,:)$ denote respectively the $\mathrm{mode-1}$, $\mathrm{mode-2}$ and $\mathrm{mode-3}$ fiber.
Let $\mathcal{T}\in \mathbb{R}^{n_1\times n_2\times m}$ a third-order tensor, $\mathcal{T}=\mathcal{X}+\mathcal{Z}$ where $\mathcal{X}$ is the signal tensor and $\mathcal{Z}$ is the noise tensor. Consider
\begin{equation}\label{pfe}
\mathcal{T} = \mathcal{X}+\mathcal{Z}=\sum_{r=1}^{q}\sigma_{r}(u_{r}^{J_1^{(r)}}\otimes w_{r}^{J_2^{(r)}}\otimes v_{r})+\mathcal{Z},\end{equation}

where $J_1^{(i)}$ and $ J_2^{(i)}$ are respectively the sets of observations indices and features indices in the $i-th$ bicluster and $u_r\in\mathbb{R}^{n_1},\;w_r\in\mathbb{R}^{n_2}$ and $ v_r\in\mathbb{R}^{m} $ are unit vectors. We assume that $u_{i}^{J_1^{(i)}}$ and $w_{i}^{J_2^{(i)}}$ have zero entries outside of ${J_1^{(i)}}$ and ${J_2^{(i)}}$ respectively for $i\in \{1,2\cdots,q\}$ and $\sigma_{1}\geq\sigma_{2}\geq\dots\geq\sigma_{q}>0$. we define $J_1~=~\bigcup_i J_1^{(i)}$ and $J_2~=~\bigcup_i J_2^{(i)}$. 
Under this model, trajectories $\mathcal{X}(J_{1},J_{2},:)$ form  at most $q$ dimensional subspace.%  where $J_1$ and $J_2$ are the indice sets of rows (observations) and columns (features) respectively, belonging to the bicluster.

Concerning  the noise model, if $(j_{1},j_{2})\notin J_{1}\times J_{2}$, we assume that entries of the noise trajectory $\mathcal{Z}(j_{1},j_{2},:)$ are independent and identically distributed (i.i.d) and each entry has a standard normal distribution. If $(j_{1},j_{2})\in J_{1}\times J_{2}$, we assume that entries of $\mathcal{Z}(j_{1},j_{2},:)$ are i.i.d and each entry has a Gaussian distribution with zero means and $\sigma_{z}^{2}$ variance. We analyse tensor biclustering problem under two variances models of the noise trajectory:
\begin{enumerate}[-]
	\item \textbf{Noise Model I:} in this model, we assume $\sigma_{z}^{2}~=~1$, i.e., the variance of the noise within and outside of the clustering is assumed to be the same. Although this model simplifies the analysis, it has the following drawback: under this noise model, for every value of $\sigma_{1}$, the average trajectory lengths in the bicluster is larger than the average trajectory lengths outside the bicluster.\\

 Indeed, let $T_1\in\mathbb{R}^{m\times k^2}$ be a matrix whose columns include trajectories $\mathcal{T}(j_1,j_2,:)$ for $(j_1,j_2)\in J_1\times J_2$ (i.e $T_1$ is the unfolded $\mathcal{T}(j_1,j_2,:)$). We can write $T_1~=~X_1 ~+~ Z_1$ where $X_1$ and $Z_1$ are unfolded $\mathcal{X}(j_1,j_2,:)$ and $\mathcal{Z}(j_1,j_2,:)$, respectively. The squared Frobinius norm of $X_1$ is equal to $\|X_1\|_F^2~=~\sigma_1^{2}$. Morever, the squared Frobenius norm of $Z_1$ has a Chi-squared distribution  with $mk^2$ degrees of freedom i.e $\chi^2(mk^2)$. Thus, the average squared Frobenius norm of $T_1$ is equal to $\sigma_1^2 + \sigma_z^2 mk^2$. Let $T_2\in\mathbb{R}^{m\times k^2}$ be a matrix whose columns include only noise trajectories. Using a similar argument, we have $\mathbb{E}[\|T_2\|_F^2]~=~mk^2$, which is smaller than $\sigma_1^2 ~+~ \sigma_z^2 mk^2$.\\
		
	\item \textbf{Noise Model II:} in this model, we assume $\sigma_{z}^{2}~=~\max\big(0, 1-(\sigma_{1}^{2}/mk^{2})\big)$, i.e., $\sigma^{2}_{z}$ is modeled to minimize the difference between average trajectory lengths within and outside the bicluster. \\
		
	Indeed, if $\sigma_1^{2} < mk^2$, without noise, the average trajectory in the bicluster is smaller than the one outside the bicluster. In this regime, having $\sigma_z^2~=~1~-~\sigma_1^2/mk^2$ makes the average trajectory lengths within and outside the bicluster comparable. This regime is called the low-SNR (signal noise ratio) regime. If $\sigma_1^2 > mk^2$, the average trajectory lengths in the bicluster is larger than the one outside the bicluster. This regime is called high-SNR regime. In this regime, adding noise to signal trajectories increases their lengths and makes solving the tensor biclustering problem easier. Therefore, in this regime we assume $\sigma_z^2~=~0$ to minimize the difference between average trajectory lengths within and outside of the bicluster.

		%If $\sigma^{2}< mk^{2}$, noise is added to make the average trajectory lengths within and outside of the bicluster comparable. 
\end{enumerate}

\subsection{Tensor Folding and Spectral (FS) }%(Fig. \ref{tfs})}
The algorithm and the asymptotic behaviour of this method are available in \cite{map}.
Under the assumption  $q=1$ and $n=|n_1|=|n_2|$, we drop the subscript $(1)$ from $J_1^{(1)}$ and $J_2^{(1)}$. We assume that $|J_1|=|J_2|=k$.  The author propose to provide only one bicluster. This method separates the selection of the two sets $J_1$ and $J_2$ using lateral slice and horizontal slice of the tensor respectively. 

\begin{eqnarray}
T_{(j_{1},1)} = \mathcal{T}(j_{1},:,:)\quad\text{ and }\quad T_{(j_{2},2)} = \mathcal{T}(:,j_{2},:) \label{t+s} \\
C_{1}=\sum_{j_{2}=1}^{n}T_{(j_{2},2)}^{t}T_{(j_{2},2)}\quad\text{ and }\quad C_{2}=\sum_{j_{1}=1}^{n}T_{(j_{1},1)}^{t}T_{(j_{1},1)} \label{c1_c2} 
\end{eqnarray} 

The aim is to select the row and column indices whose trajectories are highly correlated. The elements of $J_1$ and $J_2$ are the indices of the top $k$ elements of the top eigenvector of the matrix $C_1$ and $C_2$ respectively (algorithm \ref{algo:tensorFS}). We denoted by $\hat{J}_1$ and $\hat{J_2}$ the subset of individuals and features respectively in the bicluster given from the algorithm. % (see equation \ref{c1_c2}).
\begin{algorithm}[h!]
	\DontPrintSemicolon % Some LaTeX compilers require you to use dontprintsemicolon instead
	\KwIn{tensor $\mathcal{T}$, and the cardinality of output $k$}
	\KwOut{The set of indices $\hat{J_1}$ and  $\hat{J}_2$}
	Input: $\mathcal{T}$, $k$\;
	Initialize: $C_1$, $C_2$, $T_1$ and $T_2$\;
	\For{ i in $[n]$} {
		Compute $T_1$ according to equation (\ref{t+s})\;
		Update $C_1$ according to equation (\ref{c1_c2})\;
		Compute $T_2$ according to equation (\ref{t+s})\;
		Update $C_2$ according to equation (\ref{c1_c2})\;
	}
	Compute $\hat{u}_1$, the top eigenvector of $C_1$\;
	Compute $\hat{w}_1$, the top eigenvector of $C_2$\;
	Compute $\hat{J_1}$, set of indices of the $k$ largest values of $|\hat{u}_1|$\;
	Compute $\hat{J_2}$, set of indices of the $k$ largest values of $|\hat{w}_1|$\;
	\Return{$\hat{J_1}$ and $\hat{J}_2$}\;
		
	\caption{Tensor folding and spectral}
	\label{algo:tensorFS}
\end{algorithm}

 Tensor FS method have the best performance in both noise models compared to the three another methods (tensor unfolding+spectral, thresholding sum of squared and individual trajectory lengths) proposed by Soheil Feizi, Hamid Javadi, David Tse \cite{map}. 

\newpage
\section{Extension of Tensor Folding and Spectral}
\label{sec:extension}
%\subsection{Proposition}
In this section, we aim to extract many  biclusters in the tensor data  and improve the quality of the result. We propose several methods in order to do this task. However instead of seeking only one bicluster, we assume  that in equation \eqref{pfe} $q=r\geq 2$ where $r$ is defined by the number of gap in the eigenvalues of the covariance matrix $C_1$ and $C_2$ (equation \eqref{c1_c2}). %
%Knowing the influence of the signal strength (singular value) on the four algorithms, So there are a method which does not perform well when the gap between the singular value $\sigma_1 ? \sigma_2$ is low, for instance the thresholding individual trajectory length and sum of squared trajectory length. 
%The tensor FS perform  still a good choice (\cite{map}, \emph{section 5}). For that our contribution focus only on the method of tensor FS.
%About the tensor noise, there is no modification for the model $I$.  
\subsection{Recursive extension}
The classical method extracts one rank of the low dimensional subspace which is not very interesting because it neglect the majority of the data sets. So, Direct improvement of this method is to compute recursively according to the number of gap shown in the eigenvalues (algorithm \ref{algo:recursive}). In this method, there is no intersection in two different blocks of tensor biclustering.

%Note that a bad choice of the parameter $k$  may ruin the results, thus we propose a method to choose this parameter appropriately in the next section.

%There is only one rank of the low dimensional subspace in the classic method of tensor biclustering (algorithm \ref{algo:recursive}). In our data model, having a gap in the first and second eigenvalues guaranties the existence of the bicluster \cite{map}. 
%This method is a repetition of the classical method multiple times until the number $r$ of required bicluster is reached. Note that a bad choice of the parameter $k$  may ruin the results, thus we propose a method to choose this parameter appropriately in the next section.

%This method is based in repetition of the simple tensor FS method. After each selection of one bicluster, the entries of the trajectories selected return to zero. We use the new data set for the next bicluster and so on, until the number of required bicluster is  reached.

\begin{algorithm}
	\DontPrintSemicolon % Some LaTeX compilers require you to use \dontprintsemicolon instead
	\KwIn{tensor $\mathcal{T}$, array cardinality of each bicluster $k$}
	\KwOut{The set of all couple set of bicluster}
	$i \longleftarrow 1$\;
	\While{$i < k.length + 1$} {
		compute the first bicluster $(\hat{J}_{1}^{(i)},\;\hat{J}_{2}^{(i)})$ (by using the algorithm \ref{algo:tensorFS})\;
		keep  $(\hat{J}_{1}^{(i)},\;\hat{J}_{2}^{(i)})$ on the dataset and change the entries  to zero. We use it as a new dataset\;
		$i \longleftarrow i + 1$
	}
	\Return{$(\hat{J}_{1}^{(1)},\;\hat{J}_{2}^{(1)}),\;(\hat{J}_{1}^{(2)},\;\hat{J}_{2}^{(2)})\cdots$}\;
	\caption{ Recursive extension}
	\label{algo:recursive}
\end{algorithm}
\subsection{Multiple biclusters}

This method extract simultaneously the $r$ biclusters in our tensor by using the idea of top $r$ principal component analysis (PCA). %, where $r$ is defined by the number of significant singular values of the matrix $C_1$ and $C_2$ \eqref{c1_c2}. 
%So, before applying this method, the first step is to check the number of block available on the data set. 
The orthogonality of the principal components favor the quality of the result (algorithm \ref{algo:fsext}). 
\begin{figure}[h]
	\centering
	%\subfigure[Mode-3 fibers $\mathcal{T}(i,j,:)$]{
	\includegraphics[width=9.5cm, height=6cm]{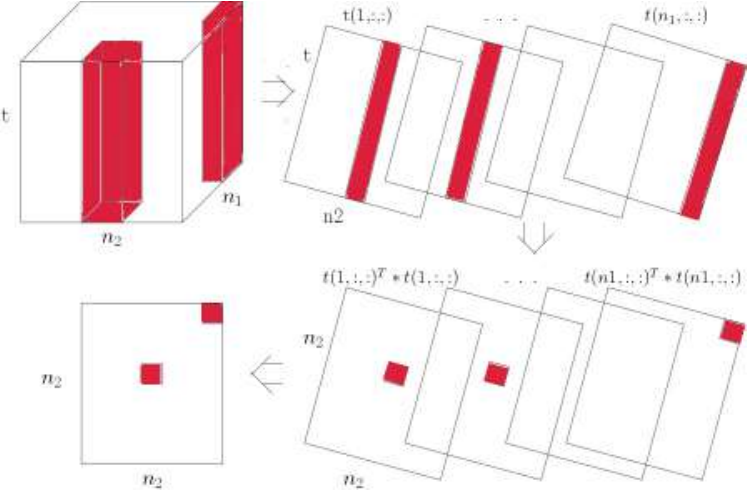}
	%	}
	\caption{A visualization of the tensor FS extension algorithm  to compute the bicluster index $(J_2^{(i)})_i$. Here we have two biclusters and the sets $J_2^{(1)}$ and $J_2^{(2)}$ do not intersect.}
	\label{tensorextension12}
\end{figure}

The illustration step of tensor FS method is showed in the Fig.\ref{tensorextension12}. For each fix individual, we have a horizontal slice of the tensor represented by $m\times n_2$ matrix (equation \eqref{t+s}). Then, we compute the covariance matrix for each horizontal slice and their sum give us only one squared matrix of order $n_2$ ($C_2$ in equation \eqref{c1_c2}). We apply singular value decomposition (SVD) in $C_2$, the top $r$ eigenvectors in the matrix $C_2$ ensure the selection of the elements of the features index set $(J_{2}^{(i)})_{i\in[r]}$ (algorithm \ref{algo:fsext}). A similar step is applied to each lateral slice of the tensor to find all the element of the index set $(J_1^{(i)})_{i\in [r]}$. \\

Since $k$ is a fix parameter, multiple bicluster method allow some trajectory belong to many blocks of tensor biclustering. We call them a boundary of bicluster. Those boundaries are very important as they belong to the intersection of all the biclusters. Thus they have all their properties.
%
%
%\newpage
\begin{algorithm}[h!]
	\DontPrintSemicolon % Some LaTeX compilers require you to use \dontprintsemicolon instead
	\KwIn{tensor $\mathcal{T}$, and the list of cardinality of the tensor biclustering $k$}
	\KwOut{The set of all couple set of bicluster}
	$r \leftarrow$ length of $k$\;
	Compute the matrices $C_1$ and $C_2$ according to equation (\ref{c1_c2})\;
	Compute the top $r$ eigenvectors of $C_1$ and $C_2$\;
	\For{$i\leftarrow 1$ \KwTo $r$}{
		Compute $\hat{J_{1}}^{(i)}$ from eigenvector $|u_{i}|$\;
		Compute $\hat{J_{2}}^{(i)}$ from eigenvector $|w_{i}|$ \;
	}
	Compute  $I_1 \longleftarrow \bigcap_i \hat{J_{1}}^{(i)}$ and $I_2 \longleftarrow \bigcap_{i} \hat{J_{2}}^{(i)}$\;
	\Return{$\big( (\hat{J}_{1}^{(i)},\;\hat{J}_{2}^{(i)}) \big)_{i\in [r]}$ and $(I_1, I_2)$}\;
	\caption{ Multiple biclusters}
	\label{algo:fsext}
\end{algorithm}

\section{Experimentation}
\label{sec:exp}
%We apply the two contributions algorithms (recursive method and the multiple tensor biclustering method) in data set. 

\subsection{Synthetic data}
We build synthetic data to evaluate the implementation of our methods. In this dataset, we have two biclusters  with signal strength $\sigma_1$ and $\sigma_2$ such that $\sigma_1 ~>~ \sigma_2$.
We assume that $v_1$ and $v_2$ are  fixed unit vectors in $\mathbb{R}^{m}$  and $v_1 ~=~ v2$. We assume also that
 $J_1^{(1)}\bigcap J_1^{(2)}=\emptyset$ and  $J_2^{(1)}\bigcap J_2^{(2)}=\emptyset$. We have $n~=~n_1~=~n_2~=~150$, $m~=~40$ and $k=|J_1^{(1)}|=|J_1^{(2)}|=|J_2^{(1)}|=|J_2^{(2)}|=30$, we assume:
\[u_1(j_1)=\left\{
\begin{array}{rcr}
1/\sqrt{k}& & \quad\text{ for } j_1\in J_1^{(1)}\\
0& & \quad\text{ if not }\\
\end{array}\right. , 
\qquad w_1(j_2)=\left\{
\begin{array}{rcr}
1/\sqrt{k}& & \quad\text{ for } j_2\in J_2^{(1)}\\
0 & &\quad\text{ if not }\\
\end{array},
\right.\]
\[u_2(j_1)=\left\{
\begin{array}{rcr}
1/\sqrt{k}& & \quad\text{ for } j_1\in J_1^{(2)}\\
0& & \quad\text{ if not }\\
\end{array}\right. , 
\qquad w_2(j_2)=\left\{
\begin{array}{rcr}
1/\sqrt{k}& & \quad\text{ for } j_2\in J_2^{(2)}\\
0 & &\quad\text{ if not }\\
\end{array}
\right.\]

\begin{figure}[!h]
	\centering
	\subfigure[Matrix $C_1$ \eqref{c1_c2}]{
		\includegraphics[width=4.2cm,height=3.5cm]{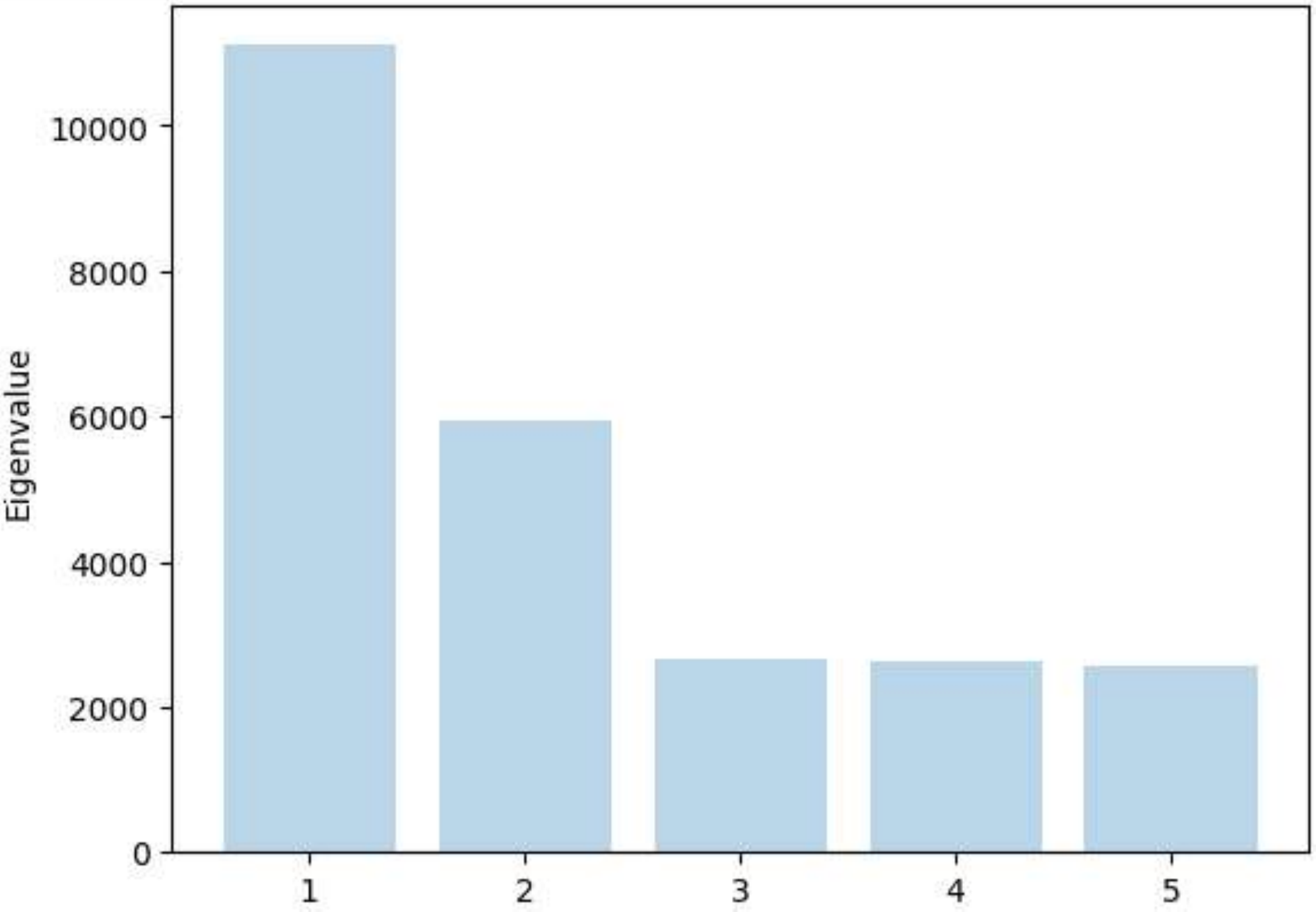}
	}
	\quad
	\subfigure[Matrix $C_2$ \eqref{c1_c2}]{
		\includegraphics[width=4.2cm,height=3.5cm]{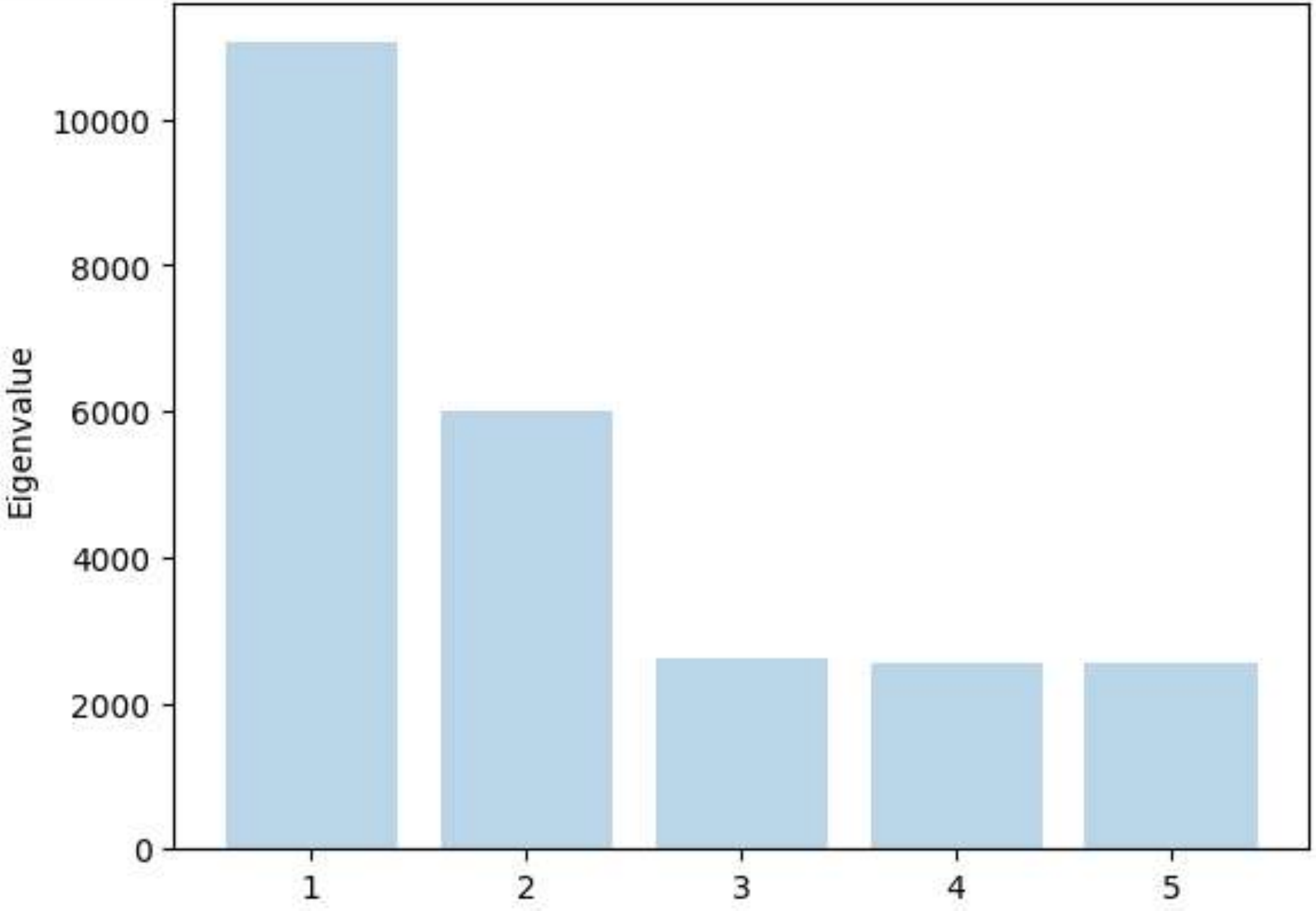}
	}\\
	\subfigure[Recovery rate with different values of $\sigma_1$.]{
		\label{evaluation}
		%of the two methods with various values of $\sigma_1$ and $\sigma_2 = 2\sigma_1/3$
		\includegraphics[width=5.5cm, height=3.7cm]{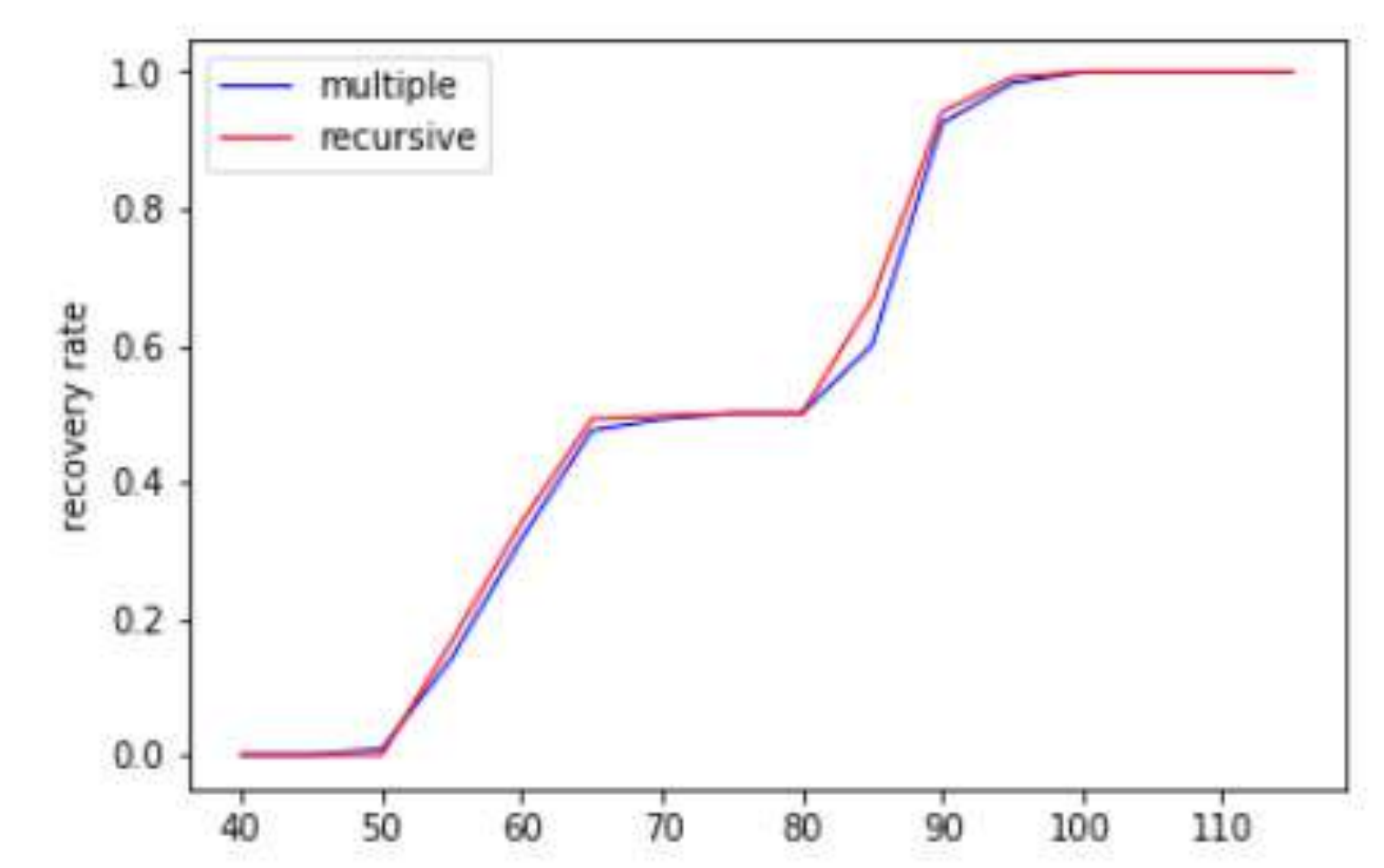}
	}	
	\caption{Synthetic data sets, $n = 150$, $m = 40$}
	\label{fig4}
\end{figure}

%\paragraph{Numerical Results:}  
We apply the assumption above to generate the input tensor $\mathcal{T}$ with the noise model II define in section 2. Let $\hat{J_1}^{(1)} \times\hat{J_2}^{(1)}$  and $\hat{J_1}^{(2)} \times\hat{J_2}^{(2)}$ be the two estimated biclusters indices of $J_1^{(1)}\times J_2^{(1)}$ and $J_1^{(2)}\times J_2^{(2)}$ respectively where $|J_1^{(1)}|=|J_2^{(1)}|=|J_1^{(2)}|=|J_2^{(2)}|=k$.
We fix the signal strength $\sigma_2 = 2\sigma_1 / 3$, if the value of $\sigma_1 > 90$, the bar plot of the top five eigenvalues of both covariance matrices tell us that there is  two block of tensor biclustering in the data (see Fig.\ref{fig4}).
%

%To define the size of the set of indices  $k$, we use the algorithm multiple tensor biclustering. The best value of $k$ is the one which gives us few intersection of the set of index in both biclusters. 

In this case, we know the value of parameter $k=30$.
To evaluate the inference quality of the result given from the algorithm, we compute the recovery rate: 
$$0\leq\frac{|\hat{J_1}^{(1)}\cap J_1^{(1)}|}{4k} +\frac{|\hat{J_1}^{(2)}\cap J_1^{(2)}|}{4k}+ \frac{|\hat{J_2}^{(1)}\cap J_2^{(1)}|}{4k} + \frac{|\hat{J_2}^{(2)}\cap J_2^{(2)}|}{4k}\leq1.$$
Recovery rate return zero if the algorithm do not find any of the element of the two biclusters and return one if the algorithm find all the elements of the two biclusters.

We did the experiment with different value of signal strength ($\sigma_1$) and for each value of $\sigma_1$ we repeat 10 times. Then we compute the average of the recovery rate (Figure \ref{evaluation}).
%\begin{figure}[!h]
%	\centering
%	\includegraphics[width=4.8cm, height=3.7cm]{evaluation_contribution.png}
%	\caption{Recovery rate of the two methods with various values of $\sigma_1$ and $\sigma_2 = 2\sigma_1/3$. }
%	\label{evaluation}
%\end{figure}
\subsection{Real Data}
\begin{figure}[t]
	%\begin{figure}[!t]
	\centering
	\subfigure[Eigenvalue $C_1$ \eqref{c1_c2}]{
		\includegraphics[width=3.4cm, height=3.5cm]{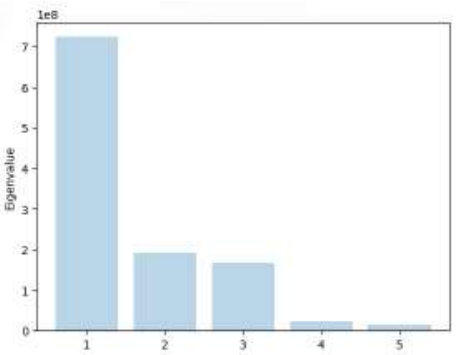}
		\label{eig_c1}
	}
	%	\caption{Real data. The top five eigenvalues of the covariance matrix.}
	%	\label{eigvalues}
	%\end{figure}
	%	\centering
	\quad
	\subfigure[First bicluster of recursive method]{
		\includegraphics[width=3.4cm, height=3.5cm]{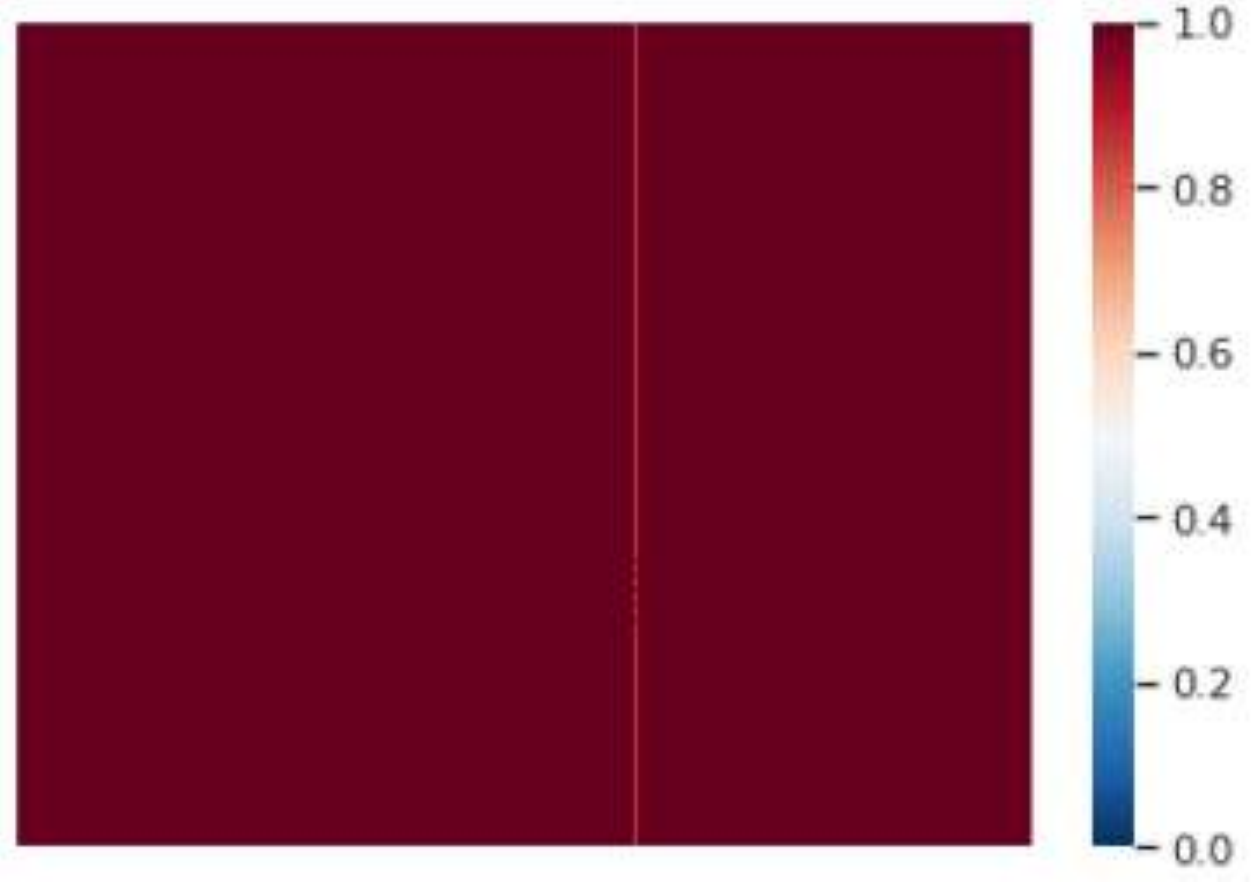}
		\label{rec1}
	}
	\quad
	\subfigure[Second bicluster of recursive method]{
		\includegraphics[width=3.4cm, height=3.5cm]{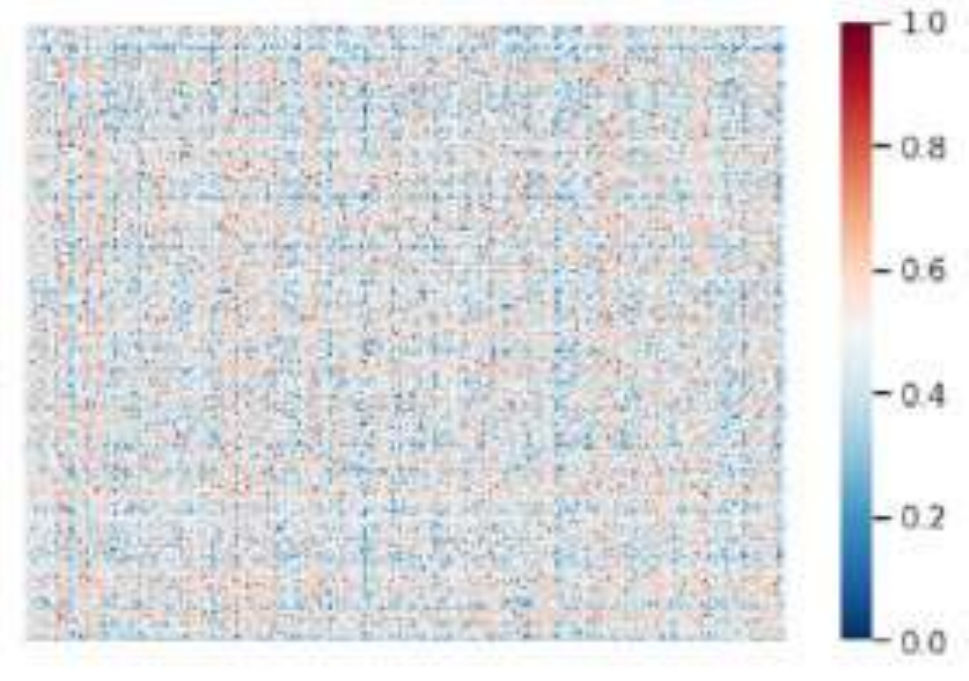}
		\label{rec2}
	}
	\quad
	\subfigure[Eigenvalue $C_2$ \eqref{c1_c2}]{
		\includegraphics[width=3.4cm, height=3.5cm]{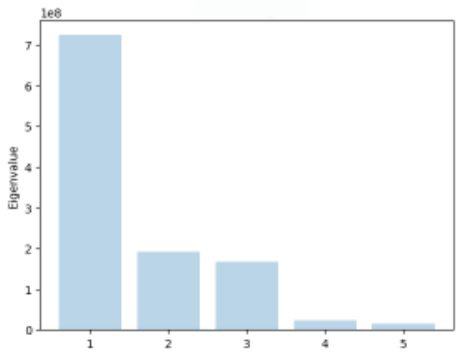}
		\label{eig_c2}
	}
	\quad
	%\end{figure}
	%\begin{figure}[t]
	\subfigure[First bicluster of multiple method]{
		\includegraphics[width=3.4cm, height=3.5cm]{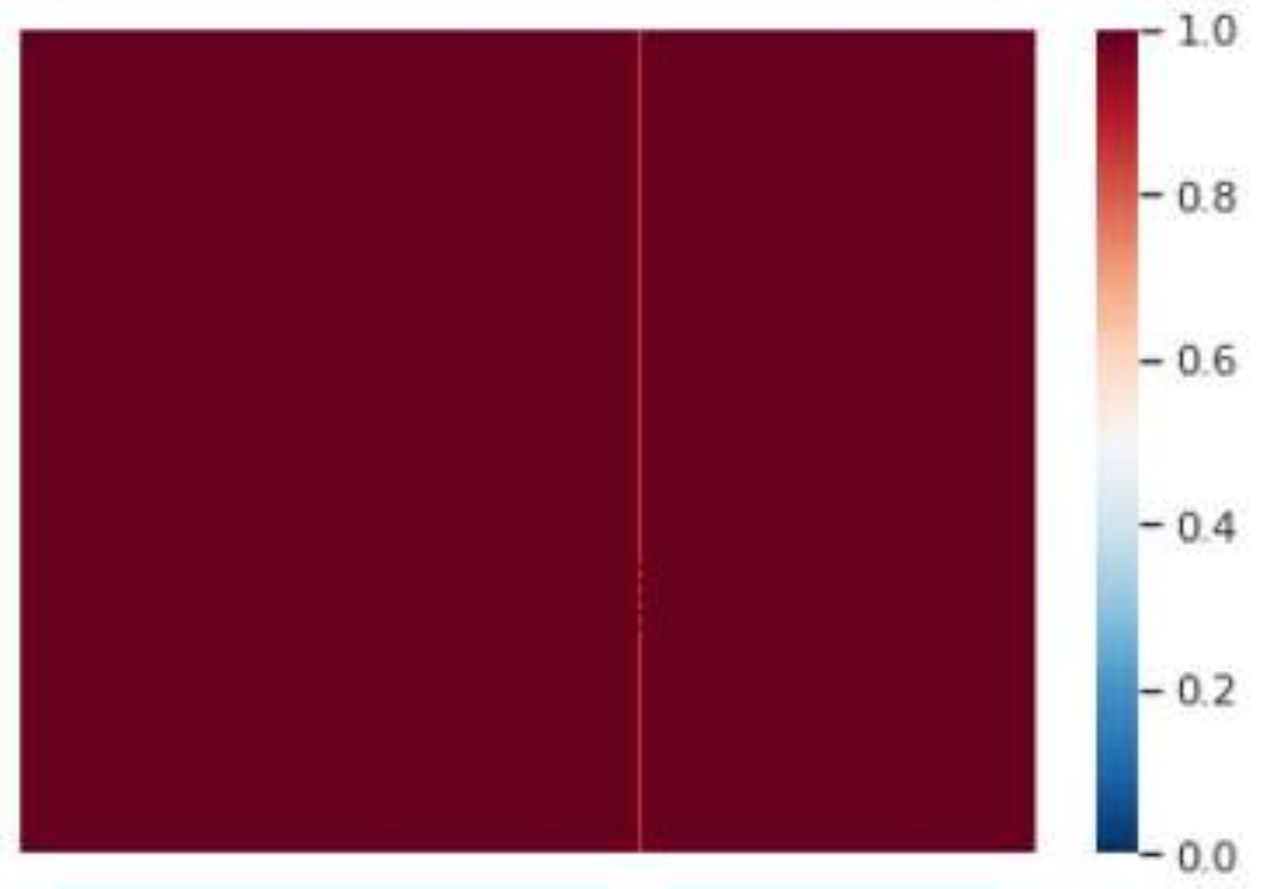}
		\label{mult1}
	}
	\quad
	\subfigure[Second bicluster of multiple method]{
		\includegraphics[width=3.4cm, height=3.5cm]{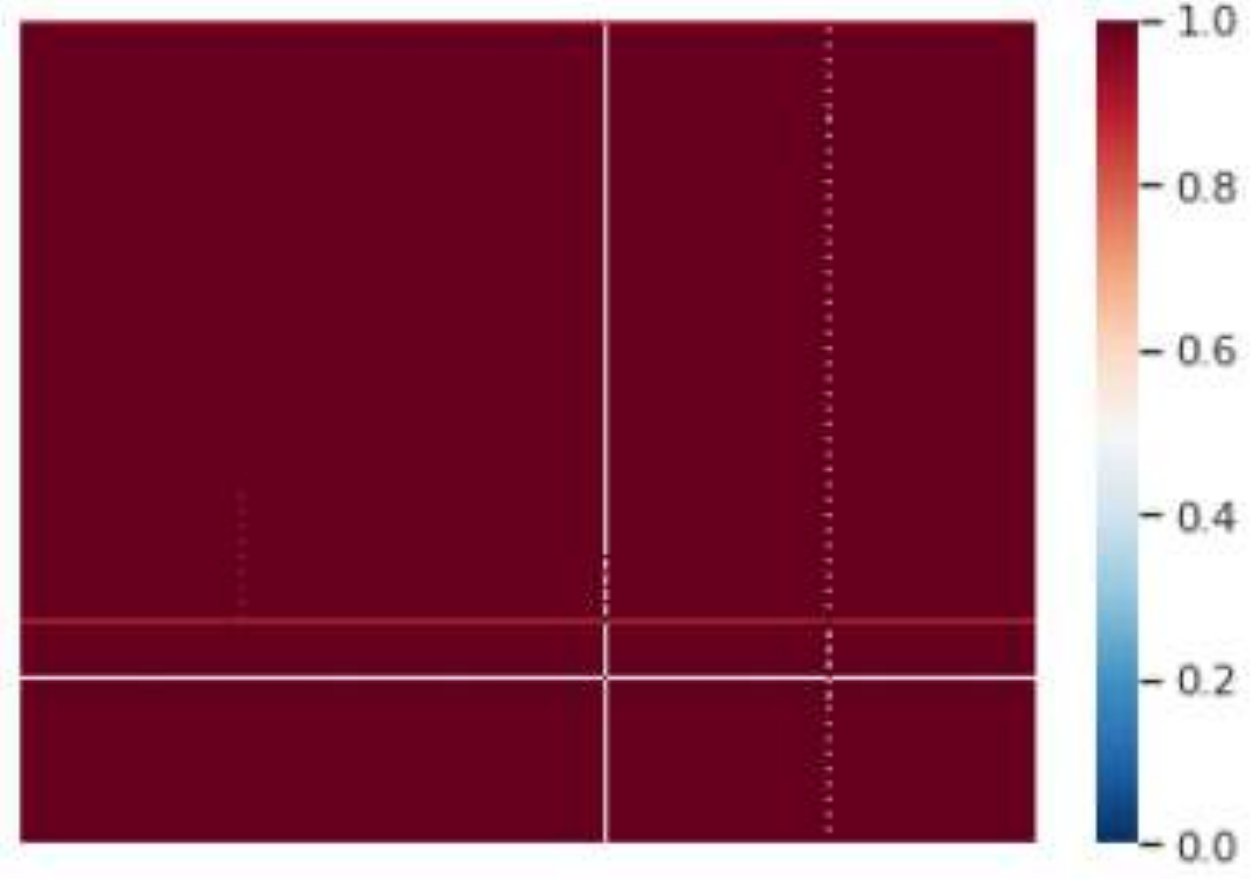}
		\label{mult2}
	}
	\caption{Trajectories correlations of each bicluster for each method}
	\label{real_result}
\end{figure}

We apply the both contribution algorithms to an electricity load diagrams \footnote{http://archive.ics.uci.edu/ml/datasets/ElectricityLoadDiagrams20112014}{data set} during four years (2011-2014). This data set contains electricity consumption of 370 clients for each 15 minutes during four years. After the data prepossessing, we have  a tensor $\mathcal{T}\in\mathbb{R}^{n_1\times n_2\times m}$ where $n_1~=~365$ is the number of day in one year, $n_2~=~161$ is the number of clients and $m~=~4$ is the number of years. \\
%
%\paragraph{Numerical Results:}
%
\indent As illustrated in Fig.(\ref{eig_c1},\ref{eig_c2}), the gap on the eigenvalues shows the existence of two tensor biclustering (section \ref{sec:extension}) in the data set. The parameter $k$ cardinality of each index sets are defined from the multiple bicluster method, we choose $k$ with few intersection of two blocks of bicluster $|J_1^{(1)}|~=~|J_1^{(2)}|~=~50$ and  $|J_2^{(1)}|~=~|J_2^{(2)}|~=~25$. % with the two contributions methods. 
After compilation, we note that  the two individuals sets $ J_1^{(1)}$ and $ J_1^{(2)}$ are disjoint in both methods. Besides,  
%there is no intersection in both methods. 
the two features sets have 22 intersection elements for multiple biclusters method and 19 intersection elements for the recursive method. So, we have two distinct  blocks with two distinct subsets of individuals and one subset of feature.\\
%one can say that the two biclusters is a two bicluster of  individuals in the same features.
%\newpage

To evaluate the quality of the bicluster given for each algorithm, we compute the total absolute pairwise correlations of the trajectories among each bicluster. With the recursive method, the the trajectories in first bicluster is highly correlated but the quality of the second bicluster is a little bit low as seen in  Figure.(\ref{rec1}, \ref{rec2}). Besides, with multiple bicluster method, the trajectories on both biclusters are highly correlated as seen in Figure.(\ref{mult1}, \ref{mult2}).
%In both methods, the trajectory of the first  bicluster $J_1^{(1)}\times J_2^{(1)}$ is very similar and highly correlated (see Fig.\ref{real_result} (c, e). In the recursive method, one trajectory can not belong to more than one bicluster and the cardinality of index set $k$ must be reached, so the pairwise correlation of the trajectory is low (see Fig.\ref{correlation} (b)). The multiple bicluster method give only information on the correlation between the trajectory and the principal component and we have a set of trajectories with high correlation (see Fig.\ref{correlation} (d)). 
%The correlation between the trajectories selected from recursive method is lower than the one selected from the multiple bicluster method (see Fig.\ref{recurcive_corr} (a,b)). 
%We see that, the multiple bicluster method perform better than the recursive method (see Fig.\ref{correlation} (b) and Fig.\ref{correlation} (d)). 
%
\section{Conclusion}
\label{sec:conc}
In this article, we introduced two methods to increase the number of bicluster selected in the tensor data set based on \cite{map}, which depends on the number of rank of the low dimensional subspace. The goal is to extract $r$ subsets of tensor ($r\geq 2$) rows and columns such that each block of the trajectories form a low dimensional subspace. We proposed  two algorithms to solve this problem, tensor recursive and multiple bicluster. 
The performance of both algorithms depends on the parameter $k$, one way to choose this parameter is in the multiple bicluster  method. If the parameter chosen gives a lot of index intersections, decreasing the value of $k$ is a good idea to improve the quality of the results. %We can change the parameter of the algorithm to a lower bound of the absolute value of the correlation between the principal component and the original vector.
%\begin{footnotesize}
\newpage

\bibliographystyle{plain}
\bibliography{bibfile}

\begin{thebibliography}{1}

\bibitem{gas}
{Andrea Montanari, Daniel Reichman and Ofer Zeitouni}.
\newblock On the limitation of spectral methods: From the gaussian hidden
  clique problem to rank-one perturbations of gaussian tensors.
\newblock {\em In Advances in Neural Information Processing Systems}, 2015.

\bibitem{bou}
Yudong Chen and Jiaming Xu.
\newblock Statistical-computational tradeoffs in planted problems and submatrix
  localization with a growing number of clusters and submatrices.
\newblock {\em arXiv preprint arXiv}, 1402, 2014.

\bibitem{fay}
Kolda and Bader.
\newblock Tensor decompositions and applications.
\newblock {\em in SIAM REVIEW}, 2009.

\bibitem{jac}
Emile Richard and Andrea Montanari.
\newblock A statistical model for tensor pca.
\newblock {\em In Advances in Neural Information Processing Systems}, 2014.

\bibitem{gud}
{Samuel B Hopkins, Jonathan Shi, and David Steurer}.
\newblock Tensor principal component analysis via sum-of-square proofs.
\newblock {\em In COLT}, 2015.

\bibitem{gol}
{Samuel B Hopkins, Tselil Schramm, Jonathan Shi and David Steurer}.
\newblock Fast spectral algorithms from sum-of-squares proofs: tensor
  decomposition and planted sparse vectors.
\newblock {\em arXiv preprint arXiv}, 2015.

\bibitem{map}
{Soheil Feizi, Hamid Javadi, David Tse}.
\newblock Tensor biclustering.
\newblock {\em Advances in Neural Information Processing Systems}, 30, 2017.

\bibitem{chr}
{T Tony Cai, Tengyuan Liang, and Alexander Rakhlin}.
\newblock Computational and statistical boundaries for submatrix localization
  in a large noisy matrix.
\newblock {\em arXiv preprint arXiv 1502.01988}, 2015.

\end{thebibliography}
%\end{footnotesize}
\end{document}